# ADLER - An efficient Hessian-based strategy for adaptive learning rate

Dario Balboni[1] and Davide Bacciu[2]

1- Scuola Normale Superiore - Data Science Department
Piazza dei Cavalieri 1, Pisa - Italy

2- University of Pisa - Computer Science Department
Largo Bruno Pontecorvo 3, Pisa - Italy

**Abstract**. We derive a sound positive semi-definite approximation of the Hessian of deep models for which Hessian-vector products are easily computable. This enables us to provide an adaptive SGD learning rate strategy based on the minimization of the local quadratic approximation, which requires just twice the computation of a single SGD run, but performs comparably with grid search on SGD learning rates on different model architectures (CNN with and without residual connections) on classification tasks. We also compare the novel approximation with the Gauss-Newton approximation.

## 1 Introduction

Recent improvements in Deep Learning have been coupled with a huge carbon footprint generated by the procedures needed to train large models [1]: for example, the common practice of employing grid search over the model hyperparameters to find the most accurate model is responsible for a multiple-fold increase in the emissions.

Hyperparameter-free optimization is a yet under-explored research direction with huge potential benefits for the development of greener deep learning, especially if applied to widely popular learning algorithms and without sacrificing performance. In this context, Stochastic Gradient Descent (SGD) is at the core of most deep learning models that are trained daily. We take a first step towards removing hyperparameters from SGD, by introducing an adaptive learning rate strategy that leverages a novel positive semi-definite (PSD) approximation to the Hessian of popular supervised optimization problems in deep learning models. Experimental results show that our strategy only requires twice as much computation as a single SGD run while showing improved performance compared to classical grid search over SGD learning rates and the Gauss-Newton method.

## 2 Algorithm Description

### 2.1 Empirical Risk Minimization in Deep Learning

Consider the typical Empirical Risk Minimization (ERM) setting in supervised learning, where we are given a set $S$ of $n$ examples $(x_i, y_i) \in X \times Y$ sampled i.i.d. from an unknown probability distribution $\mathcal{D} \in \mathcal{P}(X \times Y)$, a parametric function

$f : X \times \Theta \to Y$ and a loss function $\ell : X \times Y \times Y \to \mathbb{R}$; we are interested in the parameters $\theta \in \Theta$ that results in the smallest possible empirical loss, i.e.

$$h_S(\theta) := \frac{1}{|S|} \sum_{(x,y) \in S} \ell(x, y, f(x, \theta)).$$

To set some notation, let us define $F_S : \Theta \to Y^{\otimes n}$ as the evaluation of $f$ over all examples in $S$, i.e. $F_S(\theta) := (f(x_1, \theta), \ldots, f(x_n, \theta))$; $\mathcal{L}_S : Y^{\otimes n} \to \mathbb{R}$ by $\mathcal{L}_S(\hat{y}_1, \ldots, \hat{y}_n) := \frac{1}{n} \sum_{k=1}^{n} \ell(x_k, y_k, \hat{y}_k)$, and $h_S(\theta) := \mathcal{L}_S(F_S(\theta))$.

This view clearly separates the two functions $\mathcal{L}$ and $F$ which have different properties: $F$ can be seen as an overparameterization[1] of the "real" objective function, is randomly initialized and thus amenable to techniques related to the central limit theorem, while $\mathcal{L}$ is typically convex[2], and with its minimum value attained at $\hat{y}_k = y_k$, so that it is amenable to classical optimization techniques.

### 2.2 Proposed Hessian Approximation

It is a matter of simple calculations to rewrite the Hessian of $h_S$ as

$$\frac{\partial^2 h(\theta)}{\partial \theta_i \partial \theta_j} = \sum_{\mu\nu} \frac{\partial F_\mu(\theta)}{\partial \theta_i} \frac{\partial^2 \mathcal{L}(F(\theta))}{\partial F_\mu \partial F_\nu} \frac{\partial F_\nu(\theta)}{\partial \theta_j} + \sum_\mu \frac{\partial \mathcal{L}(F(\theta))}{\partial F_\mu} \frac{\partial^2 F_\mu(\theta)}{\partial \theta_i \partial \theta_j}$$

$$= \nabla F(\theta)^T \nabla^2 \mathcal{L}(F(\theta)) \nabla F(\theta) + \nabla \mathcal{L}(F(\theta)) \cdot \nabla^2 F(\theta),$$

where the indices $\mu, \nu$ range over the output dimensions of $F$.

Let us assume, as it commonly happens, that the last network layer is sampled independently of others from a zero mean distribution and the network output is linear in it. Multiple theoretical intuitions suggest dropping the last term:

1. For wide networks, the spectral norm of $\nabla^2 F(\theta)$ in a ball around the random initialization is $O(1/\sqrt{m})$, where $m$ is the width of the network, while the first term is $O(1)$ [2, Section 4.2, Theorem 5]. Thus the second term is a perturbation, the more negligible the wider the network is.

2. By overparameterization of the $F$ map, we expect the optimization progress to overfit the training dataset, thus ensuring that $\lim_{t \to \infty} \nabla \mathcal{L}(F(\theta_t)) = 0$[3].

3. Assume the loss function $\ell$ to be $\mathcal{D}$-symmetric in the third variable[4], i.e.

$$\forall \hat{y} \quad \mathbb{E}_{(x,y) \sim \mathcal{D}}[\ell(x, y, \hat{y})] = \mathbb{E}_{(x,y) \sim \mathcal{D}}[\ell(x, y, -\hat{y})].$$

---

[1]As the employed networks typically have more parameters than the number of examples on which they are trained, and are thus able to completely overfit the training set.

[2]This is the case for example in categorical cross-entropy and in mean squared error.

[3]This is particularly important (but it also holds more generally) as it ensures a proper recovery of the correct Hessian near the minimum, and thus enables a faithful optimization.

[4]Consider for example a regression task with MSE loss, with the target symmetric around zero. In other cases, a constant shift to the loss function ensures the argument holds.

Then, by the zero-mean random distribution of the last layer, we expect $\hat{y} = f(x, \theta)$ at initialization to be distributed randomly around zero irrespective of $x$, and thus $\mathbb{E}_{(x,y)\sim\mathcal{D}}\left[\frac{\partial}{\partial \hat{y}}\ell(x, y, f(x, \theta))\right] = 0$, and cancellations are expected to occur between the second term weights $\nabla \mathcal{L}(F(\theta))$[5].

Thus the approximation $\nabla^2 h_S(\theta) \simeq Q_S(\theta) := \nabla F_S(\theta)^T \nabla^2 \mathcal{L}_S(F_S(\theta)) \nabla F_S(\theta)$ is exact in limit settings[6], and sound in typical cases due to the expected cancellations in the second term; moreover, the approximation is PSD[7] as $\nabla^2 \mathcal{L}(F(\theta))$ is PSD because of the convexity of $\mathcal{L}$.

The presented approximation provides a Hessian that doesn't necessarily vanish as the optimization progresses, in contrast with the Gauss-Newton method[8], and enables it to better follow the local geometry of the optimization problem.

### 2.3 Per-minibatch Computation of the Learning Rate

Consider a random minibatch $I \subseteq S$ and the update equation $\theta' = \theta - \eta \nabla h_I(\theta)$: we want to search for the $\eta$ minimizing $h_I(\theta')$, that we approximate as

$$\underset{\eta}{\mathrm{argmin}}\ h_I(\theta) - \eta \nabla h_I(\theta)^T \nabla h_I(\theta) + \frac{1}{2}\eta^2 \nabla h_I(\theta)^T Q_I(\theta) \nabla h_I(\theta),$$

from which the optimal learning rate is $\eta(\theta) := \frac{\nabla h_I(\theta)^T \nabla h_I(\theta)}{\nabla h_I(\theta)^T Q_I(\theta) \nabla h_I(\theta)}$.

A difficulty in a direct application of the above formula is the high variance of the obtained estimates for the learning rate, especially problematic in later optimization phases; we thus consider an exponentially weighted average between different mini-batches approximations of the learning rate[9]. Let us call $\eta_k$ the approximation for the $k$-th minibatch; the proposed effective rate is $\hat{\eta}_k := \exp((\sum_{i=0}^{k} \beta^{k-i} \log \eta_i)/(\sum_{i=0}^{k} \beta^{k-i}))$[10], where $\beta \in [0, 1]$ is a parameter determining how much earlier estimates are considered[11].

For clarity of exposition, Algorithm 1 details the full algorithm pseudo-code.

---

[5] A more careful analysis can be made by using the Gradient Independence Assumption [3].

[6] Both on very wide networks (and NTK [4]) and on all networks at empirical convergence.

[7] It is important that the approximation is PSD as this ensures that the taken direction is a descent direction, and enables the usage of Conjugate Gradient Descent [5] on the objective.

[8] Which approximate the Hessian using $\nabla \mathcal{L}(F(\theta))^T \nabla \mathcal{L}(F(\theta))$ instead of $\nabla^2 \mathcal{L}(F(\theta))$.

[9] The underlying intuition is that gradient steps become smaller as training progresses, and thereby the local curvature becomes more stable from one minibatch to the next one, which enables a simple averaging procedure to aggregate the information contained in multiple mini-batches without a strong need to increase the mini-batch size.

[10] Since the denominator $\nabla h_I(\theta)^T Q_I(\theta) \nabla h_I(\theta)$ can be arbitrarily near zero, we need to average the *logarithms* of $\eta_k$, as this better reflects the variations expected in different estimates.

In contrast, in the Gauss-Newton approximation, we average the $\eta_k$ as they are, since in that case the denominator is with very high probability strictly greater than zero, and the distribution of the learning rates can be suitably approximated by a Gaussian distribution.

[11] Ideally, we would like to have different $\beta_k \to 1^-$, since at the start of the optimization earlier estimates aren't much informative due to large steps, while near the optimization end values of $\beta \simeq 1$ are optimal to ensure full usage of the information contained in each minibatch.

We suggest $\beta = 0.99$ as a good fixed value ensuring a decent tradeoff between being fast enough at the start and reach of the best end results; we leave $\beta$-autotuning to future research.

```
inputs
    S := {(x_i, y_i) ∈ X × Y}_{i=1}^n ;         // Input dataset
    θ_0 ∈ Θ ;                                    // Initial parameter
    b ∈ ℕ ;                                      // Size of the minibatches
    β ∈ (0, 1) = 0.99 ;      // Discount factor in exponential mean
    ε = 1e−10 ;              // Hessian Regularization Parameter
initialization: c_0 ← 0.0, s_0 ← 0.0
for k ← 0 to ∞ do
    I ← Sample(S, b) ;       // Sample minibatch from the dataset
    G ← ∇h_I(θ_k) ;          // Backward pass on the minibatch
    v ← ∇F_I(θ_k)^T G ;      // Minibatch jacobian-vector product
    Q ← ∇²L_I(F_I(θ_k)) ;    // Minibatch per-example hessian
    η ← (G^T G)/(v^T Q v + ε G^T G) ;   // Minibatch LR estimate
    s_{k+1} ← β s_k + log η ;           // LR estimate update
    c_{k+1} ← β c_k + 1.0 ;             // Denominator update
    η̂ ← exp(s_{k+1}/c_{k+1}) ;          // Effective LR
    θ_{k+1} ← θ_k − η̂ G ;               // Parameter update
end
```

**Algorithm 1:** ADaptive LEarning Rate SGD

## 3 Experimental Results

To validate the effectiveness of the proposed algorithm, we perform experiments on Convolutional architectures with and without residual connections and on Vision Transformers (ViT)[12] [6] of different widths and number of layers, different activation functions, and different random seeds, totaling more than one hundred variations[13]. For each network variation, we compare: (i) the best result of a cost-intensive grid search over ten SGD learning rates[14] ranging in logarithmic scale from 1e−5 to 3e−1; (ii) our algorithm (ADLER) with $\beta = 0.99, \varepsilon = 1e-10$; and (iii) Gauss-Newton approximation (GN) with $\beta = 0.99$.

Figure 1 and Table 1 show that ADLER performs much better than the Gauss-Newton approximation; moreover, it is on par with SGD grid search for CIFAR10 and achieves a better performance on CIFAR100. Accuracy performance apart, the method is quite effective in terms of the optimization of the given objective function, as it can be appreciated from the logarithmic slope of the training loss compared to the other methods (Figure 1).

Infact, the method is so effective at optimization that we see that the test loss starts to increase dramatically after the first ten epochs, while at the same time the accuracy performance stabilizes (Figure 1). We think that this overfitting effect is due to the employment of non-regularized ERM, and can be mitigated by a policy to increase the mini-batch size in time, which would allow the method to obtain more accurate geometrical information.

---

[12] The used networks are custom networks with residual connections, less performant than typical ResNet18/50 and are trained without batch normalization and weight decay on non-augmented datasets; this explains why the obtained results aren't SOTA.

[13] The full code and the experiment data are available at gitlab.com/dbalboni/ADLER.

[14] As our method does not have momentum terms, we only compare against naive SGD.

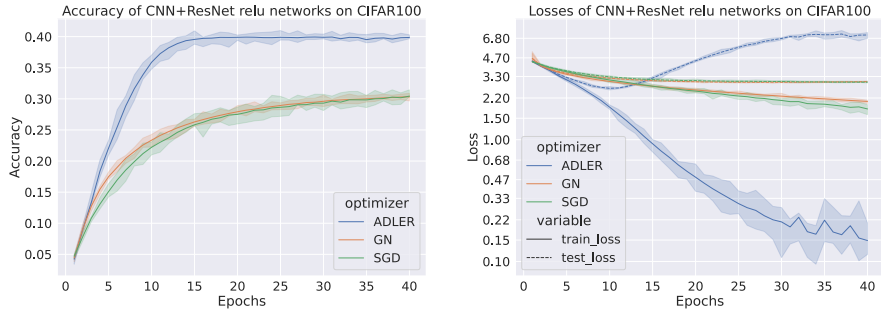

Fig. 1: Evolution of ReLU networks accuracy and training loss on CIFAR100; optimizers are shown in different colors with shaded 95% confidence intervals.

| Models | Activation | ADLER | GN | SGD |
| --- | --- | --- | --- | --- |
| CNN+ResNet | relu | **0.399 ± 0.026** | 0.303 ± 0.034 | 0.304 ± 0.056 |
| CNN+ResNet | sigmoid | 0.312 ± 0.081 | 0.357 ± 0.062 | **0.360 ± 0.022** |
| CNN+ResNet | tanh | **0.364 ± 0.040** | 0.337 ± 0.031 | 0.339 ± 0.019 |
| ViT | relu | **0.268 ± 0.007** | 0.219 ± 0.033 | 0.240 ± 0.007 |
| ViT | sigmoid | **0.270 ± 0.004** | 0.212 ± 0.042 | 0.231 ± 0.039 |
| ViT | tanh | **0.245 ± 0.024** | 0.213 ± 0.037 | 0.238 ± 0.023 |

Table 1: Final accuracy for CIFAR100 under different activations and models.

Let us now analyze the way in which the learning rates are varied throughout the learning epochs: in Figure 2 we can observe that they resemble cyclical learning rate strategies [7][15] in the case of ReLU networks; moreover we observe that the learning rates produced by GN inexorably decrease with time and are generally smaller than either of the other methods, which hints at the reason underlying its worse overall performance.

ADLER achieves a comparable accuracy to grid-search SGD on CIFAR10[16], while it shows a marked performance improvement on the more challenging CIFAR100 dataset (Table 1). While this may seem a contradiction, we think the low performance on CIFAR10 can be explained by the fact that on easier problems the local geometry appears flatter and thus the algorithm takes greater steps; these steps may sometimes be so big that the local approximation no longer holds. For this reason, we look forward to integrating a trust-region bound [8] in future extensions of the method.

---

[15] In which learning rates are increased and decreased cyclically throught training.

[16] The accuracy of the two methods is less than a standard deviation apart.

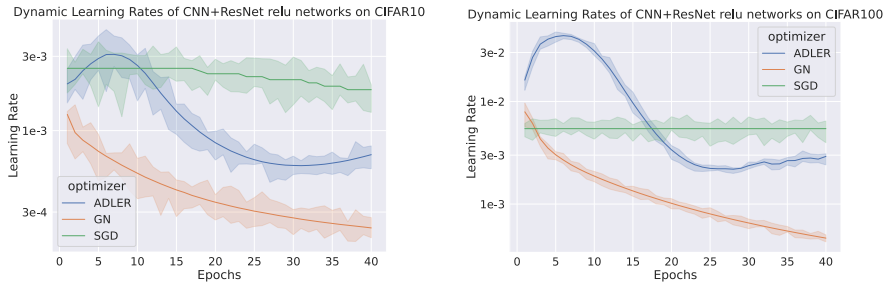

Fig. 2: Evolution of learning rates during optimization: for SGD we plot the one achieving the best performance at each epoch, while for the purpose of training, it is actually kept fixed to its grid search value.

## 4 Conclusions

In this work (1) we have presented an algorithm that relieves practitioners from computational and energy costs of grid search on SGD learning rates, without loss of predictive performance when optimizing popular convolutional architectures and Vision Transformers; (2) we have proposed a sound PSD Hessian approximation that can be easily instantiated in Conjugate Gradient methods, thus enabling to possibly match the performance of accelerated algorithms, and promoting the removal of hyperparameters from optimization procedures towards a greener AI.